\documentclass[letterpaper, 10 pt, conference]{ieeeconf}  

\IEEEoverridecommandlockouts                              

\usepackage{algorithm}
\usepackage{color}
\usepackage{tabularx}
\usepackage{multirow}
\usepackage{amsmath}
\usepackage{amssymb}
\usepackage{booktabs}
\usepackage{algpseudocode}
\usepackage{amsfonts}
\usepackage{newfloat}
\usepackage{listings}

\usepackage{times}  
\usepackage{helvet}  
\usepackage{courier}  
\usepackage[hyphens]{url}  
\usepackage{graphicx} 

\title{\LARGE \bf
Semantic-Guided Progressive Object Removal with Gaussian Splatting
}
\author{Xianliang Huang$^{{1,3}}$, Chen Xiao$^{3}$, Yuanxiang Ni$^{2}$, Guanming Liu$^{3}$, \\ Mingkai Liu $^{1}$, Dikai Fan $^{1}$, Xiao Liu$^{1}$, Hao Zhang$^{2,\dagger}$ 
\thanks{$^{1}$PICO, ByteDance Inc., $^{2}$Shenzhen Institute of Advanced Technology, Chinese Academy of Sciences, $^{3}$ Fudan University}%
\thanks{$^\dagger$For correspondence and questions: {h.zhang10@siat.ac.cn}}
}

\begin{document}

\maketitle
\thispagestyle{empty}
\pagestyle{empty}

\begin{abstract}
Removing unwanted objects from reconstructed 3D scenes is an important task in computer vision, supporting applications in AR/VR, robotics, and digital content creation. Existing methods typically complete the entire masked region in a single step and without effectively utilizing semantic information from other views, leading to difficulties in handling complex geometric details and textures. In this work, we propose a novel framework that integrates Semantic-guided Block Matching (SBM) and Region-Wise Progressive Refinement (RPR) for high-quality 3D object removal. First, we leverage DINOv2 to encode semantic guidance from multi-view observations, and the best match tokens are decoded to complete missing regions in the target view while maintaining cross-view consistency. Second, we introduce a RPR strategy that segments the target mask into multiple subregions and selectively refines those with poor visual quality. Our method is built upon Gaussian Splatting, ensuring high-fidelity scene reconstruction with efficient computation. Experimental results demonstrate that our approach outperforms existing Gaussian-based methods in terms of perceptual quality and coherence in 3D object removal.
\end{abstract}

\section{INTRODUCTION}
Three-dimensional scene reconstruction and manipulation have been significantly advanced by Neural Radiance Fields~\cite{mildenhall2021nerf} and 3D Gaussian Splatting~\cite{kerbl20233d} (3DGS), which enable photorealistic and efficient rendering for a wide range of applications, including virtual and augmented reality~\cite{mao2024live}, robotics~\cite{liu2025mace,zhang2025mind}, and autonomous driving~\cite{zhou2024drivinggaussian}. 
A fundamental yet challenging task in this domain is 3D object removal, which involves eliminating unwanted objects from scenes and realistically completing the resulting holes. 
This task becomes particularly difficult when removing large objects in unbounded 360° environments, where it is necessary to leverage multi-view observations, hallucinate previously unseen content, and maintain both visual consistency and geometric plausibility across all views. 
Among recent advances, 3DGS has emerged as a powerful solution for real-time novel view synthesis and editable 3D scene reconstruction. Consequently, accurate and consistent object removal within Gaussian-based representations is becoming increasingly crucial for interactive editing and downstream scene understanding tasks.

Despite recent advancements~\cite{mirzaei2022spin,chen2024gaussianeditor,huang2026nerf}, existing 3D object removal methods still face challenges when dealing with complex occlusions and fine-grained geometry. A key limitation lies in their insufficient exploitation of semantic information across multiple views. For instance, methods like SPIn-NeRF~\cite{mirzaei2022spin} perform inpainting primarily from 2D inputs while largely neglecting cross-view semantic consistency. As a result, they often produce inconsistent reconstructions, lacking geometric coherence in object regions.
Other approaches~\cite{chen2024mvip,liu2024infusion,zhong2025generative} leverage generative priors through Score Distillation Sampling (SDS)~\cite{poole2022dreamfusion} to optimize 3D representations. However, these approaches frequently yield visually inaccurate or overly smooth reconstructions, as they lack explicit geometric guidance and struggle to preserve high-frequency details. Furthermore, these techniques adopt a one-shot completion strategy for the entire masked region, which restricts their ability to iteratively refine suboptimal regions and correct localized artifacts.

To overcome the above limitations, we propose a novel framework that incorporates Semantic-guided Block Matching (SBM) across different views, enabling accurate recovery of missing structures and textures by aligning semantically relevant content. 
In addition, we introduce a High-frequency feature extraction module that provides auxiliary supervision to guide the generative process toward sharper inpainting results. Specifically, the High-frequency prior is injected into a pre-trained diffusion model, allowing it to synthesize plausible content conditioned on both global semantics and fine-grained visual details.

To further enhance the visual fidelity and consistency of the removal regions, we present a RPR strategy by segmenting the removal area into several blocks and selectively refining the regions with low perceptual fidelity.
Specifically, the RPR strategy is driven by frequency-aware UNet, which focuses computational resources on challenging regions while avoiding redundant updates to already satisfactory areas. This target-level refinement significantly boosts the realism and coherence in the reconstructed 3D scene, making our approach more robust and perceptually superior to existing one-shot pipelines.

Overall, our pipeline enables precise and efficient 3D object removal by addressing the core problems of multi-view generative inpainting. We integrate SBM with the RPR strategy, allowing for better structural alignment and localized detail enhancement. Extensive experiments on various datasets, which include both forward-facing and unbounded scenes, demonstrate that our framework outperforms the existing baselines in terms of visual fidelity and geometric consistency.
The key contributions of our framework are summarized as follows:
(1) We propose a novel 3D object removal framework that combines 3D Gaussian Splatting with Score Distillation Sampling to produce initial inpainting results. A High-frequency feature extractor is adopted to guide the pre-trained diffusion model in repainting sharper regions. (2) A semantic-guided block matching is introduced to enhance missing structures and textures by explicitly aligning semantically relevant content from other perspectives. (3) We propose a RPR strategy that divides the inpainting region into multiple subregions and selectively refines low-quality subregions, improving local fidelity and overall coherence of the removal region. (4) Extensive experiments are conducted to validate the effectiveness of our method, demonstrating superior performance over existing NeRF-based and Gaussian-based inpainting baselines in visual quality and geometric accuracy.

\section{Related Work}
\subsection{Diffusion Model for Conditional Generation}
Diffusion models have emerged as a powerful framework for conditional content generation, with applications spanning image restoration~\cite{zhou2024seeing}, editing~\cite{zhou2025devil}, and 3D scene understanding~\cite{ho2020denoising,huang2023iddr}. Early models such as DDPM~\cite{ho2020denoising} and Stable Diffusion~\cite{rombach2022high} demonstrated strong performance in high-fidelity image generation, while latent diffusion greatly improved efficiency and enabled conditioning via text or spatial priors~\cite{zhou2025learning}.
To enable fine-grained control over the generation process, a wide variety of conditioning techniques have been proposed. Prompt-based editing~\cite{lin2024text}, CLIP-guided modulation~\cite{zhang2023adding}, and attention modification have been used to direct generation semantically. Spatial control approaches such as SpaText~\cite{avrahami2023spatext} introduce mask-based conditioning for localized edits, while personalization methods like DreamBooth~\cite{ruiz2023dreambooth} and Textual Inversion~\cite{gal2022image} finetune diffusion models with limited user-provided data for instance-level control. For structure-aware editing tasks such as inpainting, models like SDEdit~\cite{meng2021sdedit} and DDIM Inversion~\cite{dhariwal2021diffusion,song2020denoising} inject noise into images and iteratively denoise to allow semantic modifications while preserving global structure. These techniques enable faithful reconstruction and content preservation, making them particularly effective for realistic 3D object removal scenarios.

\subsection{3D Inpainting meets Gaussian Splatting}
Early approaches for 3D inpainting primarily focused on geometric completion alone~\cite{park2019deepsdf,wang2017shape}. Recently, several works have utilized CLIP-based optimization~\cite{gao2023textdeformer,wang2022clip}, Score Distillation Sampling~\cite{poole2022dreamfusion,zhuang2023dreameditor}, and Iterative Dataset Update strategies~\cite{huang2024hi} to distill the 2D generative prior into 3D inpainting techniques. These techniques are commonly applied across different representations, including meshes~\cite{michel2022text2mesh}, NeRFs~\cite{weber2024nerfiller,barda2024instant3dit}, and SDFs~\cite{cheng2023progressive3d}.
With the development of 3DGS~\cite{kerbl20233d}, 
Gaussian-based 3D inpainting methods~\cite{liu2024infusion,chen2024gaussianeditor,ye2025gaussian} have emerged as a promising path in filling missing regions within the GS framework, taking advantage of its superior rendering efficiency and high-quality reconstruction.
GScream~\cite{wang2024learning} enhances information flow between visible and occluded areas, promoting both geometric consistency and texture coherence in complex regions. MVInPainter~\cite{cao2024mvinpainter} leverages reference-guided multi-view inpainting, significantly simplifying in-the-wild novel view synthesis by utilizing unmasked clues instead of explicit pose inputs.
GaussianEditor~\cite{chen2024gaussianeditor} introduces Hierarchical Gaussian Splatting and Gaussian semantic tracing to improve the precision and stability of diffusion-guided reconstructions. Infusion~\cite{liu2024infusion} performs 3D inpainting by learning depth completion from diffusion priors, while Gaussian Grouping~\cite{ye2025gaussian} often suffers from inaccurate unseen region segmentation during mask generation, adversely affecting inpainting quality.
In this work, we further advance 3D inpainting within the Gaussian Splatting framework by improving both visual fidelity and computational efficiency.

\begin{figure*}[ht]
\centering\includegraphics[width=1\textwidth]{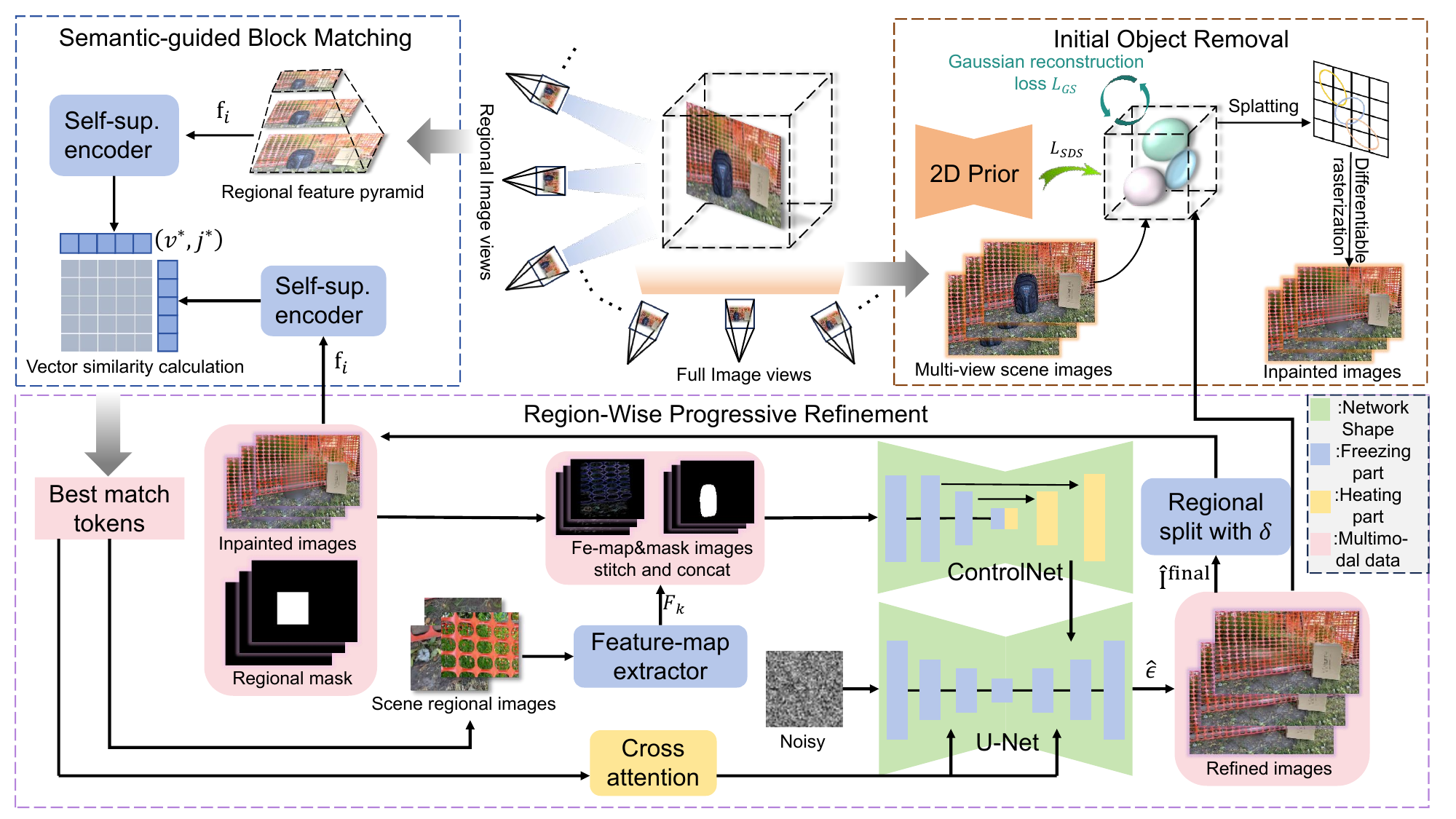}
\caption{\textbf{Overview of our pipeline.} The key insight of our method is to iteratively refine all region blocks in a few steps, guided by the high-frequency information from the semantic-guided block. Specifically, the semantic-guided block matching leverages multi-view observations to guide the completion of missing regions with content-aligned semantics. Next, the high-frequency feature extractor provides auxiliary supervision signals from image components to enhance detail preservation during diffusion-based generation. Finally, we utilize a RPR strategy that decomposes the target mask into subregions and iteratively refines visually suboptimal areas for improved local and global coherence.
}
\label{fig:pipeline}
\vspace{-0.4cm}
\end{figure*}

\section{Preliminary}
\subsection{Gaussian Splatting}
We use 3DGS~\cite{kerbl20233d} to model 3D scenes, representing them as anisotropic Gaussian primitives. Each primitive $G_i = ({x_i}, {r_i}, \alpha_i, {c_i})$ includes a center ${x_i} \in \mathbb{R}^3$, rotation ${r_i} \in \mathbb{R}^3$, opacity $\alpha_i \in [0, 1]$, and RGB color ${c_i} \in \mathbb{R}^3$ via spherical harmonics. The Gaussian density is:
\begin{equation}
    G(\mathbf{x}) = e^{-\frac{1}{2}(\mathbf{x} - \mathbf{\mu})^T \Sigma^{-1} (\mathbf{x} - \mathbf{\mu})},
\end{equation}
with covariance $\Sigma = \mathbf{R} \mathbf{S} \mathbf{S}^T \mathbf{R}^T$, where $\mathbf{R}$ is rotation and $\mathbf{S}$ is scaling.
For rendering, 3D Gaussians are projected to 2D using the camera matrix, processed in parallel as pixel blocks~\cite{zwicker2001ewa}. The final color $\hat{\mathbf{C}}$ is computed via alpha blending~\cite{max1995optical}:
\begin{equation}
    \hat{\mathbf{C}} = \sum_{i=1}^{N} {c}_i \cdot \alpha_i \cdot \prod_{j=1}^{i-1}(1 - \alpha_j),
\end{equation}
where ${c}_i$ and $\alpha_i$ are the color and opacity of the $i$-th Gaussian, ordered by depth.


\subsection{2D Diffusion Models}
Diffusion models operate via two key processes: a forward diffusion step $q(\mathbf{z}_\tau | \mathbf{z}_0)$ that incrementally corrupts a data sample $\mathbf{z}_0 \sim p_\text{data}(\mathbf{z})$ with Gaussian noise, and a learned reverse process that gradually denoises a pure Gaussian sample $\mathbf{z}_T \sim \mathcal{N}(0, \mathbf{I})$ back to a clean image. At a particular timestep $\tau \in [0, T]$, the noisy latent $\mathbf{z}_\tau$ can be expressed as:
\begin{equation}
\mathbf{z}_\tau = \sqrt{\bar{\alpha}_\tau} \, \mathbf{z}_0 + \sqrt{1 - \bar{\alpha}_\tau} \, \boldsymbol{\epsilon}, 
\end{equation}
where $\boldsymbol{\epsilon} \sim \mathcal{N}(0, \mathbf{I})$ denotes the injected noise and $\bar{\alpha}_\tau$ is the cumulative product of noise schedule coefficients.
This forward process transitions into the reverse diffusion phase, where a neural network \(\boldsymbol{\epsilon}_\theta\) learns to predict the noise \(\hat{\boldsymbol{\epsilon}} = \boldsymbol{\epsilon}_\theta(\mathbf{z}_\tau, \tau, \mathbf{c})\), starting from a pure Gaussian sample \(\mathbf{z}_T \sim \mathcal{N}(0, \mathbf{I})\). The training objective minimizes the denoising error 
\begin{equation}
\mathcal{L}_\text{diff} = \left\| \boldsymbol{\epsilon} - \hat{\boldsymbol{\epsilon}} \right\|^2,
\end{equation}
and using this predicted noise, the denoised latent at the previous timestep is computed as \(\mathbf{z}_{\tau-1} = \mathbf{z}_\tau - \hat{\boldsymbol{\epsilon}}\) (with scaling omitted for brevity). This procedure is repeated iteratively to reconstruct a clean sample \(\mathbf{z}_0\). For latent diffusion models such as Stable Diffusion~\cite{rombach2022high}, $\mathbf{z}$ resides in a compressed latent space where \(\mathbf{z} = \mathcal{E}(\mathbf{x})\) maps pixels to latents and a decoder \(\mathcal{D}(\cdot)\) generates the final image \(\mathbf{x}\).


\subsection{Diffusion Priors for 3D Generation} 
Beyond 2D synthesis, diffusion models serve as powerful generative priors for optimizing implicit 3D representations like NeRF~\cite{mildenhall2021nerf} and 3DGS~\cite{kerbl20233d}, unlike GANs or deterministic models~\cite{suvorov2022resolution}. In this work, Score Distillation Sampling (SDS)~\cite{poole2022dreamfusion} is utilized to update Gaussian primitives \(G_\theta\) such that the rendered image $\mathbf{x}_r$ aligns with the distribution of real images under the learned diffusion model. By adding noise \(\boldsymbol{\epsilon}\) to obtain \(\mathbf{z}_\tau\), and minimizing:
\begin{equation}
\mathcal{L}_\text{SDS} = \left\| \boldsymbol{\epsilon} - \boldsymbol{\epsilon}_\theta(\mathbf{z}_\tau, \tau, \mathbf{c}) \right\|^2.
\end{equation}
We backpropagate this gradient between the predicted and actual noise to refine \(\theta\), guiding the 3DGS to produce photorealistic renderings consistent with 2D diffusion prior.

\section{Methodology}
In this section, we detail the design of SBM, the High-frequency Feature Extractor and RPR in Fig.~\ref{fig:pipeline}. Our goal is to achieve high-fidelity 3D object removal with cross-view consistency and fine-grained visual quality. 



\subsection{Multi-view Representation and Initialization}
Formally, 3D scenes can be represented by 3D Gaussians $\theta$, given a collection of multi-view images $\mathcal{I} = \{I_i\}^n_{i=1}$, accompanied by respective camera poses $\{\pi_i\}^n_{i=1}$. 
To initialize the 3D Gaussian set, we first estimate a sparse point cloud from Structure-from-Motion (SfM)~\cite{schonberger2016structure}. The initial positions $\boldsymbol{x_i}$ of the Gaussians are seeded from these points, while their scales and opacities are set to default or random values. 
To increase scene fidelity, we perform adaptive densification through primitive splitting and cloning strategies, dynamically increasing the density of Gaussians in regions with high reconstruction error.

The 2D object masks $\mathcal{M} = \{M_i\}^n_{i=1}$ are utilized to indicate the target 3D object.
We remove the Gaussians in the masked region according to object masks and replace them with the same amount of randomly initialized Gaussians. Then, the Score Distillation Sampling~\cite{poole2022dreamfusion} loss is calculated by distilling knowledge from the inpainting backbone~\cite{rombach2022high} to improve the rendering result of 3DGS with multi-step noise prediction. The initial inpainted images $\hat{\mathcal{I}} = \{\hat{I}_i\}^n_{i=1}$ are formulated as:
\begin{equation}
\begin{aligned}
                \hat{I}_k =  G_{\theta}({I_k}, \pi_k),
 \end{aligned}
\end{equation}
where $G_\theta$ denotes the 3DGS model, $I_k$ is the input view, and $\pi_k$ is its corresponding camera pose.

To update the parameters $\theta$, Gaussian noise $\epsilon$ is added to the rendered images $\hat{\mathbf{I}}$ and the following SDS loss is applied, using predicted noise $\hat{\epsilon}_\psi$ from the latent diffusion model $\psi$:
\begin{equation}
\begin{aligned}
\nabla_\theta \mathcal{L}_{\text{SDS}}(\psi, \hat{\mathbf{I}}) = \mathbb{E}_{{\epsilon},t}\left[ w(t) \big(\hat{\epsilon}_\psi(\hat{\mathbf{I}}_t; y,t) - {\epsilon}\big)\frac{\partial \hat{\mathbf{I}}}{\partial \theta} \right],
 \end{aligned}
\end{equation}
where $y$ would be the input text and $t$ is the level of noise.

\subsection{Semantic-Guided Block Matching}
\label{sec:block-matching}
To fully utilize cross-view contextual cues, we propose a \textit{semantic-guided block matching strategy} to optimize the ambiguity of target regions by retrieving semantically matched content across views and enforcing high-level alignment through diffusion-based generation. It serves as a key component to improve multi-view consistency and fine-grained realism in removal regions.

\subsubsection{Block Decomposition and Semantic Extraction}
The initial inpainted image $\hat{\mathbf{I}} \in \mathbb{R}^{H \times W \times 3}$ with its corresponding binary mask $\mathbf{M} \in \{0,1\}^{H \times W}$ indicating the removal object, we first divide both $\mathbf{I}$ and $\mathbf{M}$ into $N$ non-overlapping square blocks $\mathbf{B}_i$ with size $s \times s$.
Only blocks sufficiently occluded, satisfying $\mathbf{B}_i \cap \mathbf{M} > \tau$, are selected as target blocks for retrieving.  The occlusion ratio threshold $\tau$ is empirically set to 0.1 in our experiments. Specifically, a smaller $\tau$ increases the number of selected blocks, capturing more occluded areas but reducing performance due to higher computational demands. Conversely, a larger $\tau$ selects fewer blocks, improving performance but potentially lowering accuracy by ignoring moderately occluded blocks.

For each block $\mathbf{B}_i$, we extract high-level semantic embeddings using a pre-trained DINOv2 encoder $\mathcal{F}_{\text{DINOv2}}$:
\begin{equation}
    \mathbf{f}_i = \mathcal{F}_{\text{DINOv2}}(\mathbf{B}_i \odot M) \in \mathbb{R}^{d},
\end{equation}
where $d$ is the feature dimension. 

\subsubsection{Cross-View Retrieval}
The original images $\{\mathbf{I}^{v}\}_{v=1}^{V}$ partitioned into blocks $\{\mathbf{B}^{v}_j\}$, we compute semantic features $\mathbf{f}^{v}_j = \mathcal{F}_{\text{sem}}(\mathbf{B}^{v}_j)$ for all visible blocks.
For each target block $\mathbf{B}^{t}_i$, we perform semantic matching by identifying the best matching source block $\mathbf{B}^{v^*}_{j^*}$ via cosine similarity:
\begin{equation}
    (v^*, j^*) = \arg\max_{v,j} \, \text{sim}(\mathbf{f}^{t}_i, \mathbf{f}^{v}_j),
\end{equation}
where $\text{sim}(\cdot, \cdot)$ denotes cosine similarity between two feature vectors.

\subsubsection{Semantic-guided Inpainting}
The core idea of this part is to encode the selected blocks that are located with relevant features from other viewpoints and inject these features into a pre-trained diffusion model to refine the inpainted images.
Once matched, the selected source block $\mathbf{B}^{v^*}_{j^*}$ is used as a semantic reference to guide the denoising process of the diffusion model~\cite{rombach2022high} for generating the corresponding object region. Specifically, the matched source block $\mathbf{B}^{v^*}_{j^*}$ is projected into the latent space, where probabilistic sampling is performed using a UNet-based denoising network. To incorporate semantic priors, we replace the original text embedding $\mathbf{c}$ with a semantic-aware token $\mathbf{c}_i$, derived from the matched reference block. These tokens are injected into each layer of the UNet via cross-attention mechanisms, enabling fine-grained, semantic-driven conditioning.

Formally, let $\mathbf{z}^{t}_i$ denote the latent of the occluded target block at timestep T. The predicted noise is computed as:
\begin{equation}
    \hat{\boldsymbol{\epsilon}} = \boldsymbol{\epsilon}_{\theta}(\mathbf{z}^{t}_i, T, \mathbf{c}_i).
\end{equation}
This guidance mechanism ensures that the generated content within each masked block is semantically aligned with the most relevant information from other views, resulting in more coherent and visually plausible inpainting results.

\begin{figure}[t]
	\centering
	\scalebox{1}{\includegraphics[width=1.0\linewidth]{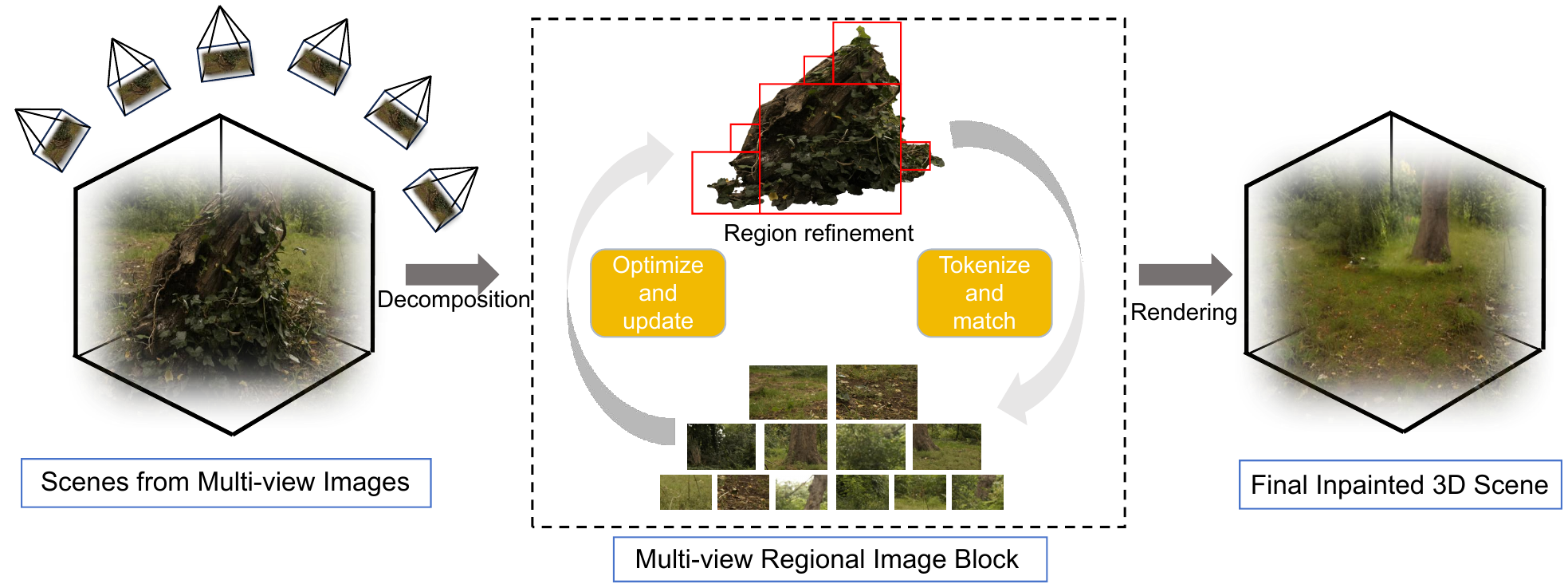}}
	
	\caption{
		  Given a 3D scene, our method produces accurate object removal with region refinement and outputs a visually consistent 3D scene.
	}
	 \label{fig:teaser}
     \vspace{-0.4cm}
\end{figure}

\subsection{Region-Wise Progressive Refinement Strategy}
\label{sec:refinement}
Although semantic alignment guides the initial inpainting, the results may still exhibit blurry textures or geometric inconsistencies, especially in complex regions. To address this, we introduce an RPR strategy, which selectively refines low-quality blocks using a frequency-aware diffusion generator. The RPR process is illustrated in Fig.~\ref{fig:teaser}.


\subsubsection{High-Frequency Feature Extraction}
We first identify low-quality inpainted regions based on their lack of High-frequency details. Specifically, we adopt our High-frequency Feature Extractor, which is inspired by edge detection and combines horizontal and vertical gradient responses of the grayscale image. Given a reconstructed RGB image $\mathbf{I} \in \mathbb{R}^{H \times W \times 3}$, we first convert it to grayscale $\mathbf{I}_{\text{gray}}$, then apply high-pass filtering using horizontal and vertical Sobel kernels~\cite{kanopoulos1988design} $\mathbf{K}_h$ and $\mathbf{K}_v$, respectively. The high-frequency response $\mathbf{F}_k$ is computed as:
\begin{equation}
    \mathbf{F}_k = (\mathbf{I}_{\text{gray}} \otimes \mathbf{K}_h + \mathbf{I}_{\text{gray}} \otimes \mathbf{K}_v) \odot \mathbf{I} \odot \mathbf{M}_{\text{erode}},
    \label{eq:high}
\end{equation}
where $\otimes$ denotes convolution and $\odot$ is the Hadamard product. $\mathbf{M}_{\text{erode}}$ is an eroded version of the original mask to exclude noisy boundaries. A block is marked for refinement if its average high-frequency magnitude is below a predefined threshold $\delta$.
A block is labeled “low-quality” if its high-frequency amplitude is below 15\% of the surrounding area, a threshold chosen empirically based on frequency statistics across scenes. 

\subsubsection{Re-Inpainting with Frequency Guidance}
For each $\mathbf{B}_k \in \mathcal{B}_{\text{low}}$, we re-infer the block using a diffusion process based on the pre-trained UNet. Unlike the initial inpainting step, here we concat both semantic token guidance and a frequency-aware embedding to better preserve high-frequency textures.

Given a latent representation $\mathbf{z}_k$ and its semantic token $\mathbf{c}_k$ obtained via DINOv2 matching, we also encode the block’s high-frequency map $\mathbf{F}_k$ into a compact feature vector $\mathbf{f}_k$ using a lightweight convolutional encoder $\mathcal{E}_{\text{freq}}$:
\begin{equation}
\mathbf{f}_k = \mathcal{E}_{\text{freq}}(\mathbf{F}_k), \quad \mathbf{F}_k = |\nabla_x \mathbf{I}_k| + |\nabla_y \mathbf{I}_k|,
\end{equation}
where $\nabla_x$ and $\nabla_y$ denote Sobel-filter gradients in the horizontal and vertical directions, respectively.

We then condition the denoising UNet on both $\mathbf{c}_k$ and $\mathbf{f}_k$:
\begin{equation}
\hat{\boldsymbol{\epsilon}} = \boldsymbol{\epsilon}_{\theta'}(\mathbf{z}_k, T, \mathbf{c}_k, \mathbf{f}_k),
\end{equation}
where $\boldsymbol{\epsilon}_{\theta'}$ denotes the fine-tuned UNet, and T is the diffusion timestep. The frequency embedding $\mathbf{f}_k$ is injected through a cross-attention mechanism or MLP fusion block into intermediate UNet layers, enabling the network to recover sharper textures guided by both global semantics and local detail priors.


After generating the refined result $\hat{\mathbf{B}}_k$ for each low-quality block $\mathbf{B}_k \in \mathcal{B}_{\text{low}}$, we selectively replace the corresponding region in the initially completed image:
\begin{equation}
\hat{\mathbf{I}}^{\text{final}} = \text{Replace}(\hat{\mathbf{I}}, \mathbf{B}_k \leftarrow \hat{\mathbf{B}}_k), \quad \forall \mathbf{B}_k \in \mathcal{B}_{\text{low}}.
\end{equation}
This progressive and targeted replacement strategy ensures that only perceptually suboptimal regions are refined, thereby reducing unnecessary computation.


\begin{figure*}[ht]
	\centering
 \includegraphics[width=1.0\linewidth]{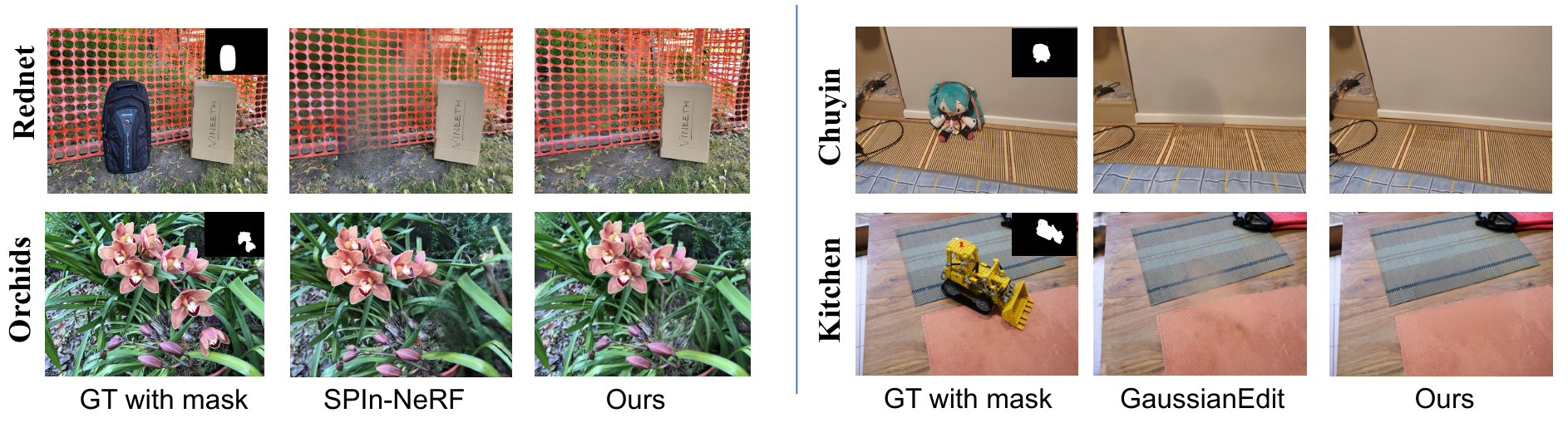}
	\caption{
		\textbf{Qualitative results.} We compare our method with SPIn-NeRF and GaussianEdit in a consistent view.
	}
	\label{fig:compare_with_nerf}
\end{figure*}

\subsection{Integration with Gaussian Splatting}
Our object removal framework is ultimately integrated into a 3DGS pipeline to enable consistent reflection in the underlying 3D geometry and appearance. 
The Gaussian representation is updated by optimizing both color and spatial parameters across views, guided by a composite loss function that promotes both local realism and global multi-view consistency.
The total loss combines the Gaussian reconstruction loss with SDS loss terms, weighted to balance multi-view coherence and visual consistency.

Gaussian reconstruction loss combines a mean absolute error (MAE) term and a differentiable SSIM loss~\cite{wang2004image}:
\begin{equation}
\mathcal{L}_\text{GS} = (1 - \lambda)\mathcal{L}_\text{1} + \lambda \mathcal{L}_\text{D-SSIM},
\end{equation}
where $\mathcal{L}_1 = \|\hat{\mathbf{I}} - \mathbf{I}\|_1$ measures pixel-level differences, and $\mathcal{L}_\text{D-SSIM}$ emphasizes perceptual structure similarity. The balance factor $\lambda$ is consistent with the original 3DGS.

The overall loss function used to supervise the optimization of the Gaussian parameters $\theta$ is:
\begin{equation}
    \mathcal{L}_{\text{total}} = \mathcal{L}_\text{GS} + \lambda_{\text{SDS}} \mathcal{L}_{\text{SDS}},
\end{equation}
where $\lambda_{\text{SDS}}$ is a hyperparameter that control the influence of semantic-aware guidance.
To maintain efficiency, we limit the number of Gaussians and employ a coarse-to-fine strategy, initializing with sparse Gaussians and progressively refining them. As a result, our method enables realistic object removal, high-fidelity novel view synthesis, and reliable downstream scene understanding within an efficient and lightweight 3DGS framework.

\section{Experiments}
In this section, we evaluate our proposed approach for 3D object removal in several benchmark datasets. We first describe our implementation details, then introduce the datasets, baselines, and evaluation metrics. We present both quantitative and qualitative results, followed by ablation studies to analyze the contribution of block-matching strategy and progressive refinement strategy.

\subsection{Experimental Setup}
\subsubsection{Implementation Details}
Our method is implemented on top of the official 3D Gaussian Splatting (3DGS) PyTorch CUDA extension. 
To ensure stable convergence, we apply Gaussian densification and pruning between iterations 100 and 2500. The 3D scene is initialized using sparse point clouds obtained via Structure-from-Motion (SfM), and progressively refined throughout the training.
All experiments are conducted on an NVIDIA RTX 3090 GPU (24GB memory). We adopt the Adam optimizer with an initial learning rate of $1 \times 10^{-4}$ and $s$ is set to 15.
Following the SDS strategy from DreamFusion~\cite{poole2022dreamfusion}, we perform 1000 iterations of Score Distillation Sampling to obtain an initial diffusion prior. During this process, latent features are perturbed using a pre-defined noise schedule, and a frozen conditional Imagen model predicts the added noise to guide parameter updates.
We employ a pre-trained Variational Autoencoder to encode both the original RGB image and its binary mask into a 4-channel latent space, serving as the input to the diffusion model. Stable Diffusion v2.1~\cite{rombach2022high} is adopted as the base generator throughout all experiments.

\begin{table*}[ht]
    \centering
        \caption{Quantitative comparisons with Gaussian-based methods on SPIn-NeRF, Self-captured, and Mip-NeRF 360 datasets.}
    \begin{tabular}{l|ccc|ccc|ccc}
        \toprule
        \multirow{2}{*}{Method} & \multicolumn{3}{c|}{\textit{SPIn-NeRF dataset}} & \multicolumn{3}{c|}{\textit{Self-captured dataset}} & \multicolumn{3}{c}{\textit{Mip-NeRF 360 dataset}} \\
        & PSNR ↑ & SSIM ↑ & LPIPS ↓ & PSNR ↑ & SSIM ↑ & LPIPS ↓ & PSNR ↑ & SSIM ↑ & LPIPS ↓ \\
        \midrule
        SPIn-NeRF & 26.8 & 0.901 & 0.176 & 21.7 & 0.824 & 0.231 & 23.0 & 0.860 & 0.204 \\
        GaussianEdit & 22.5 & 0.841 & 0.217 & 25.2 & 0.879 & 0.192 & 24.9 & 0.871 & 0.208 \\
        \toprule
        GaussianGroup & 21.7 & 0.815 & 0.174 & 19.1 & 0.693 & 0.188 & 20.8 & 0.684 & 0.193 \\
        MVInpainter & 27.5 & 0.915 & 0.160 & 26.1 & 0.891 & 0.179 & 25.8 & 0.884 & 0.191 \\
        InFusion & 27.1 & 0.912 & 0.164 & 26.4 & 0.893 & 0.174 & 25.5 & 0.881 & 0.186 \\
        \toprule
        \textbf{Ours} & \textbf{28.7} & \textbf{0.929} & \textbf{0.139} & \textbf{29.6} & \textbf{0.936} & \textbf{0.131} & \textbf{27.4} & \textbf{0.899} & \textbf{0.161} \\
        \bottomrule
    \end{tabular}

    \label{tab:gaussian_quantitative}
    \vspace{-0.4cm}
\end{table*}

\subsubsection{Datasets and Baselines} To comprehensively evaluate the effectiveness of our proposed method, we conduct experiments on both public datasets~\cite{mildenhall2019llff,barron2022mip,mirzaei2022spin} and self-captured datasets, covering a wide range of scene complexities and camera configurations. 
We compare our method against several state-of-the-art approaches for 3D scene object removal, including both NeRF-based and Gaussian-based methods. 

\subsubsection{Evaluation Metrics}
To quantitatively evaluate the effectiveness of our method, we adopt Peak Signal-to-Noise Ratio (PSNR), Structural Similarity Index Measure (SSIM)~\cite{wang2004image}, and Learned Perceptual Image Patch Similarity (LPIPS)~\cite{zhang2018unreasonable} to quantitatively assess synthesis results.
These metrics are calculated solely for the pixels within removal regions identified in each view to evaluate the image quality.

\subsection{Performance Comparisons}
\subsubsection{Results on Forward-facing Scenes} We compare our method against two representative 3D inpainting approaches: SPIn-NeRF~\cite{mirzaei2022spin} and GaussianEdit~\cite{chen2024gaussianeditor}. The evaluated scenes are performed on \emph{Rednet} from SPIn-NeRF, \emph{Orchids} from LLFF, \emph{Chuyin} from self-captured dataset and \emph{Kitchen} from Mip-NeRF. 
We assess the performance in terms of maintaining scene coherence and visual quality. In the left part of Fig.~\ref{fig:compare_with_nerf}, SPIn-NeRF struggles with texture discontinuities and blurry regions in texture complex areas, such as the geometric structure of grass and net in \emph{Rednet} and \emph{Orchids}. 
For the right part scene of \emph{Chuyin} and \emph{Kitchen},
GaussianEdit demonstrates slightly better texture fidelity but often suffers from ghosting artifacts due to its reliance on the removal of Gaussian primitives.
In contrast, our method leverages multi-view semantic alignment and high-frequency supervision to produce sharper and semantically plausible results. 
Tab.~\ref{tab:gaussian_quantitative} reports PSNR, SSIM, and LPIPS metrics computed over masked inpainting regions. 
Our method consistently outperforms the SPIn-NeRF and GaussianEdit across all metrics, indicating superior reconstruction quality and perceptual realism.

\subsubsection{Comparison with Gaussian-based Methods} 
\begin{figure*}
	\centering
	\scalebox{1}{\includegraphics[width=1.0\linewidth]{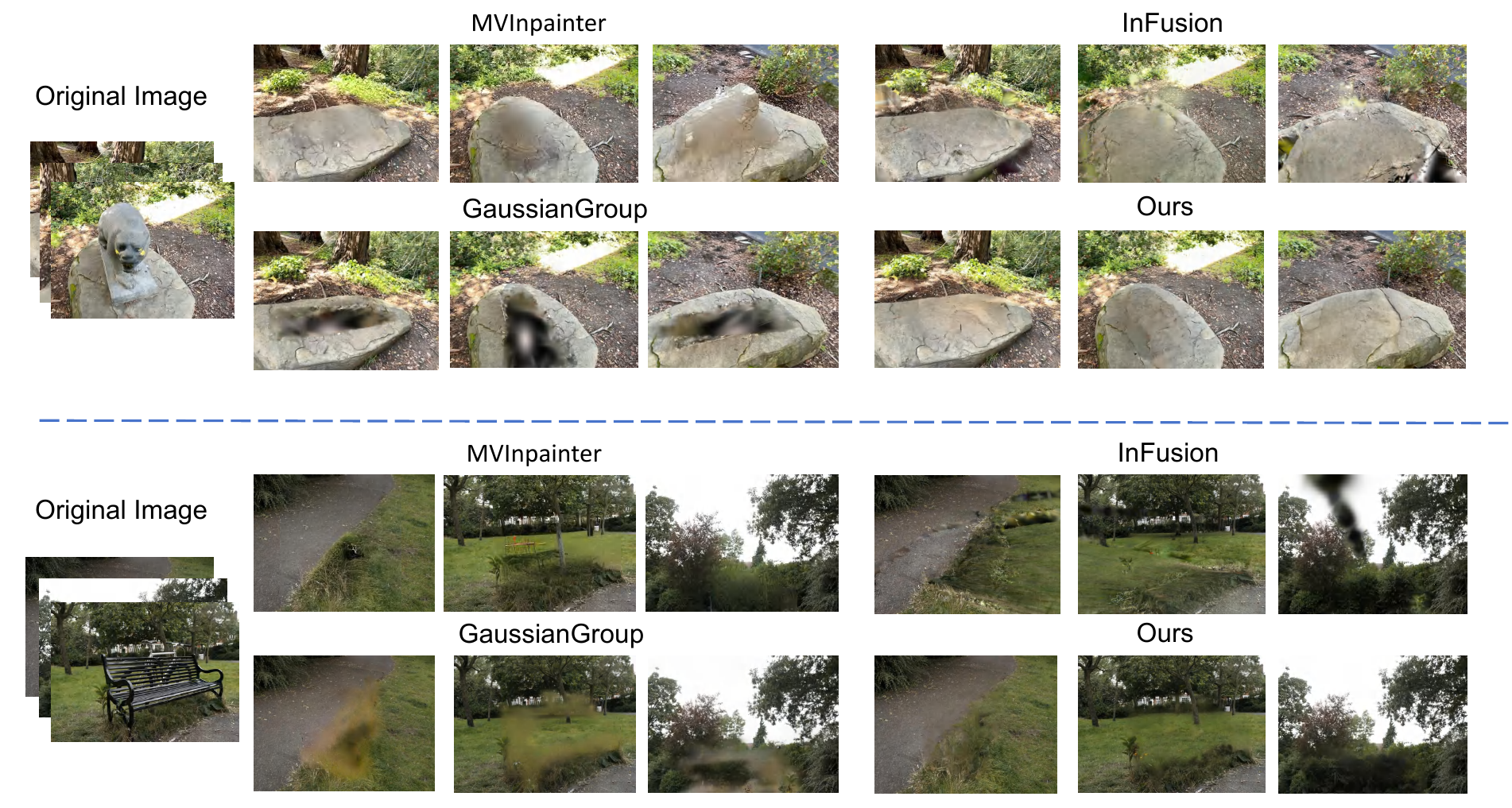}}
	
	\caption{
		 \textbf{Qualitative results on \emph{Bear} and \emph{Bicycle} scenes.} Given the input images, we visualize the removal results of MVInpainter, GaussianGroup, InFusion and our method in a consistent view. 
	}
	\label{fig:gaussian_comparison}
\vspace{-0.4cm}
\end{figure*}

We further evaluate our method against recent state-of-the-art Gaussian-based methods, including {MVInpainter}~\cite{cao2024mvinpainter}, {GaussianGroup}~\cite{ye2025gaussian}, and {InFusion}~\cite{liu2024infusion}. 
These methods leverage 3D Gaussian Splatting representations to enable photorealistic object removal. Our comparisons focus on multi-view consistency across challenging scenes. 
As shown in Fig.~\ref{fig:gaussian_comparison}, our method delivers cleaner and more semantically aligned inpainting in the \textit{Bear} and \textit{Bicycle} scenes. 
MVInpainter produces visually smooth results but struggles with detailed restoration and sometimes introduces over-smoothed patches.
GaussianGroup introduces semantically irrelevant textures or severely inconsistent shading, particularly in highly cluttered backgrounds.  InFusion achieves promising results guided by 2D diffusion priors but lacks precise structural alignment in object removal areas. In contrast, our method benefits from semantic-aware token guidance and progressive region-wise refinement, achieving visually realistic and structurally coherent results across views.
Tab.~\ref{tab:gaussian_quantitative} presents a quantitative comparison of our proposed method with existing Gaussian-based approaches across PSNR, SSIM, and LPIPS metrics computed on masked regions. Our approach achieves the highest performance across all evaluation metrics, demonstrating its effectiveness in 3D object removal.
Specifically, for the \textit{Rednet} and \textit{Chuyin} scenes, our method achieves a peak PSNR of 28.7, significantly outperforming other baselines. 
In the more challenging \textit{Kitchen} scene, our approach continues to outperform all competitors, with a PSNR of 27.4, SSIM of 0.899, and LPIPS of 0.162—outclassing the previous best method InFusion.
Overall, our method consistently surpasses other baselines across all scenes and metrics, highlighting its strengths in perceptual similarity, structural fidelity, and view-consistent content completion.

\subsubsection{Comparison on Runtime and Memory} We compare the training time, inference time, and GPU memory usage of various methods on our self-captured dataset using an NVIDIA RTX 3090 GPU. As shown in Tab.~\ref{tab:efficiency}, our method achieves a favorable trade-off between performance and efficiency.
Specifically, MVInpainter and InFusion require 20GB memory and longer training durations , while GaussianEditor and GaussianGroup offer faster training but with limited accuracy. In contrast, our method requires only 7.9 hours of training and 13 minutes of inference time, with a modest GPU memory footprint of 10GB, outperforming most baselines in efficiency without compromising quality.
These results demonstrate the practicality of our approach for real-world deployment scenarios, where both runtime and memory are critical considerations.

\begin{table}[t]
\centering
\caption{\textbf{Efficiency Comparison.} We report training time, inference time, and GPU memory consumption.}
\footnotesize
\setlength{\tabcolsep}{3pt}
\newcolumntype{Y}{>{\centering\arraybackslash}X}
\begin{tabularx}{\linewidth}{l|YYY}
\toprule
\textbf{Method} & \textbf{Training Time} & \textbf{Inference Time} & \textbf{GPU Memory} \\
\midrule
\scriptsize SPIn-NeRF        & 20h   & 5h     & -     \\
\scriptsize GaussianEditor   & 2h    & 10min  & 12GB  \\
\scriptsize GaussianGroup    & 2.6h  & 20min  & 15GB  \\
\scriptsize MVInpainter      & 120h  & 5min   & 20GB  \\
\scriptsize InFusion         & 24h   & 2min   & 20GB  \\
\scriptsize \textbf{Ours}    & 7.9h  & 13min  & 10GB \\
\bottomrule
\end{tabularx}
\label{tab:efficiency}
\vspace{-0.4cm}
\end{table}

\subsection{Ablation Studies}
SBM module enhances geometric alignment across views by leveraging semantic features to guide correspondence during multi-view inpainting. As shown in Tab.~\ref{tab:ablation}, removing SBM causes a substantial degradation in performance, with the PSNR dropping by 5.57 and 4.44 on the self-captured and public scenes, respectively. SSIM also decreases notably, while LPIPS increases by 0.186 and 0.218, indicating a significant loss in perceptual quality. 
RPR module progressively refines inpainting results from coarse to fine across semantic regions, ensuring detail preservation and boundary smoothness. 
As shown in Tab.~\ref{tab:ablation}, when RPR is removed, PSNR drops by 0.78 and 2.19, while LPIPS increases by 0.017 and 0.047 for the self-captured and public scenes, respectively. Although the SSIM degradation is less pronounced, the increased LPIPS reflects reduced perceptual realism. 
These results highlight the importance of both SBM and RPR, which improves boundary fidelity.

\begin{table}[t]
\centering
\caption{\textbf{Ablation Study.} We report the average Metrics on the Self-captured scene and the Public scene.}
\footnotesize
\setlength{\tabcolsep}{3pt}
\newcolumntype{Y}{>{\centering\arraybackslash}X}
\begin{tabularx}{\linewidth}{l||YYY|YYY}
\toprule
\multirow{2}{*}{\textbf{Method}} & \multicolumn{3}{c|}{\textit{Self-captured scene}} & \multicolumn{3}{c}{\textit{Public scene}} \\
& PSNR$\uparrow$ & SSIM$\uparrow$ & LPIPS$\downarrow$ & PSNR$\uparrow$ & SSIM$\uparrow$ & LPIPS$\downarrow$ \\
\midrule
w/o SBM              & 24.58 & 0.681 & 0.312 & 24.18 & 0.605 & 0.369 \\
w/o RPR              & 29.37 & 0.761 & 0.143 & 26.43 & 0.841 & 0.198 \\
\midrule
\textbf{Full Model}  & \textbf{30.15} & \textbf{0.886} & \textbf{0.126} & \textbf{28.62} & \textbf{0.892} & \textbf{0.151} \\
\bottomrule
\end{tabularx}
\label{tab:ablation}
\vspace{-0.4cm}
\end{table}

\section{Conclusion}
\label{sec:conclusion}
We propose a novel framework for 3D object removal and scene completion that combines semantic-guided inpainting and RPR within 3DGS pipeline. By leveraging multi-view semantic features extracted via DINOv2, our method performs block-level matching to guide the completion of occluded regions while preserving cross-view consistency. A progressive refinement strategy further enhances visual quality by selectively updating low-fidelity regions based on perceptual feedback. Experiments show that our method outperforms existing Gaussian-based baselines in perceptual quality.

\vspace{-0.1cm}
\section{Acknowledgement}
This project was supported by the National Natural Science Foundation of China under Grant No.~62472415.

\bibliographystyle{IEEEtran.bst}
\bibliography{reference}









\end{document}